\documentclass[10pt,twocolumn,letterpaper]{article}

\usepackage[pagenumbers]{cvpr} 
\usepackage{multirow}
\usepackage{graphicx}
\usepackage{amsmath}
\usepackage{amssymb}
\usepackage{booktabs}
\usepackage{multirow}

\usepackage[pagebackref,breaklinks,colorlinks]{hyperref}

\usepackage[capitalize]{cleveref}
\crefname{section}{Sec.}{Secs.}
\Crefname{section}{Section}{Sections}
\Crefname{table}{Table}{Tables}
\crefname{table}{Tab.}{Tabs.}

\def\cvprPaperID{*****} 
\def\confName{CVPR}
\def\confYear{2022}

\begin{document}

\title{Performance degradation of ImageNet trained models by simple image transformations}

\author{Harsh Maheshwari\\
Georgia Institute of Technology\\
{\tt\small harshmaheshwari@gatech.edu}
\and
}
\maketitle

\begin{abstract}
ImageNet trained PyTorch models are generally preferred as the off-the-shelf models for direct use or for initialisation in most computer vision tasks. In this paper, we simply test a representative set of these convolution and transformer based models under many simple image transformations like horizontal shifting, vertical shifting, scaling, rotation, presence of Gaussian noise, cutout, horizontal flip and vertical flip and report the performance drop caused by such transformations. We find that even simple transformations like rotating the image by $10^{\circ}$ or zooming in by 20\% can reduce the top-1 accuracy of models like ResNet152 by 1\%+. The code is available at \href{https://github.com/harshm121/imagenet-transformation-degradation}{https://github.com/harshm121/imagenet-transformation-degradation}
\end{abstract}

\section{Introduction}
\label{sec:introduction}

Classification performance on ImageNet dataset is being constantly pushed forward \cite{papersithcode}. On many computer vision tasks (including image classification on ImageNet) we find new papers with an increment in performance on the task by a few \%. However, do we know how significant is this difference of a few \% in accuracy?

The aim of this report is to evaluate the standard PyTorch image classification models under various simple image transformations. The report tries to bring forward that if such simple transformations like rotating the image by just $10^{\circ}$ can lead to a drop in accuracy by more than 1\%, is an increment of a few \% in accuracies significant enough? 

Moreover, some of these transformations, like scaling, translation, random occlusions can very well occur in real life and it is critical to know, understand and acknowledge the performance degradation of these models on such transformations.

ImageNet trained models have been critized before for not being able to generalize well on new datasets \cite{generalization-imagenet, ObjectNet}, but we don't aim to evaluate these models on new datasets, which are sometimes testing the limits of these models. Rather, we focus on simple image transformations on the original validation set of ImageNet and report the Top-1 and Top-5 accuracies.

There has been an active interest \cite{ObjectNet, imagenet-a, imagenet-rendition, imagenet-sketch, imagenetv2} in the machine learning community to check the robustness and generalization capabilities of image classification models on out-of-distribution datasets. Specifically, \cite{ObjectNet} varies the object backgrounds, rotations, and imaging viewpoints randomly and aim to expose the biases that might have creeped in by training on the standard ImageNet dataset. \cite{imagenet-rendition, imagenet-sketch} create datasets with a slight twist by finding art-clips and sketches of the classes present in ImageNet and check if the standard models are able to predict their classes. \cite{imagenet-a} creates two even more challenging datasets by an adversarial filtration technique aiming to fool the models. All these existing works go one step ahead and create datasets to check the degradation in performance. Instead of testing on images sampled from elsewhere, we on the other hand, investigate the performance of the ImageNet trained models on the same images, but with simple transformations. Slight scaling of objects ideally should not lead to any degradation in performance, as when these models are deployed in other settings, it is very likely that the scale of the objects during inference might be different from the usual scale of the ImageNet data. Similarly, during inference, the objects might also not be present in the center of the image, which is mostly the case in the ImageNet dataset, so if the performance degrades just by slightly translating the images, it is a cause of concern.

Even though these transformations are simple, we found that some of them led to significant drop in performance. For example, scaling the images to just 0.8x led to 1.4\% drop and rotating the image by just $5^{\circ}$ led to a drop by 0.48\% in ResNet-152's top-1 accuracy. We vary the severity of such transformations and provide all the results in Section \ref{sec:results} and Appendix \ref{sec:numbers}

\section{Method}
\label{sec:method}
We use convolution models ResNet-50 and ResNet-152 and ViT models ViT-B/16 and ViT-L/16 loaded with ImageNet trained weights off the shelf from \cite{pytorch} and  \url{https://github.com/lukemelas/PyTorch-Pretrained-ViT}. All the results are reported on the $50k$ images in the validation set of ImageNet dataset.

Transformations were performed using the \textit{imgaug} \cite{imgaug} library. We chose rotation (5, 10, 25, 45, 90 degrees clockwise and anti-clockwise), additive Gaussian noise (standard deviation varying between 0 and 0.05, 0.1, 0.2, 0.3, 0.4 $\times$ 255), rectangular cutout (of sizes 0.05, 0.1, 0.2, 0.3, 0.4, 0.5 $\times$ height and width), horizontal translation (2\%, 5\%, 10\%, 15\%, 25\%, 35\%, 50\% left and right), vertical translation (2\%, 5\%, 10\%, 15\%, 25\%, 35\%, 50\% up and down), scaling the image (to 0.2, 0.3, 0.4, 0.5, 0.6, 0.7, 0.8, 0.9, 1.1, 1.2, 1.3, 1.4, 1.5, 1.6, 1.75 $\times$ the height and width) and flipping the image (left-right and up-down). The images were transformed and then passed to the model to evaluate the Top1 and Top5 classification accuracies.
\section{Results}
\label{sec:results}
\begin{figure}[!h]
\centering
\includegraphics[width=0.45\textwidth]{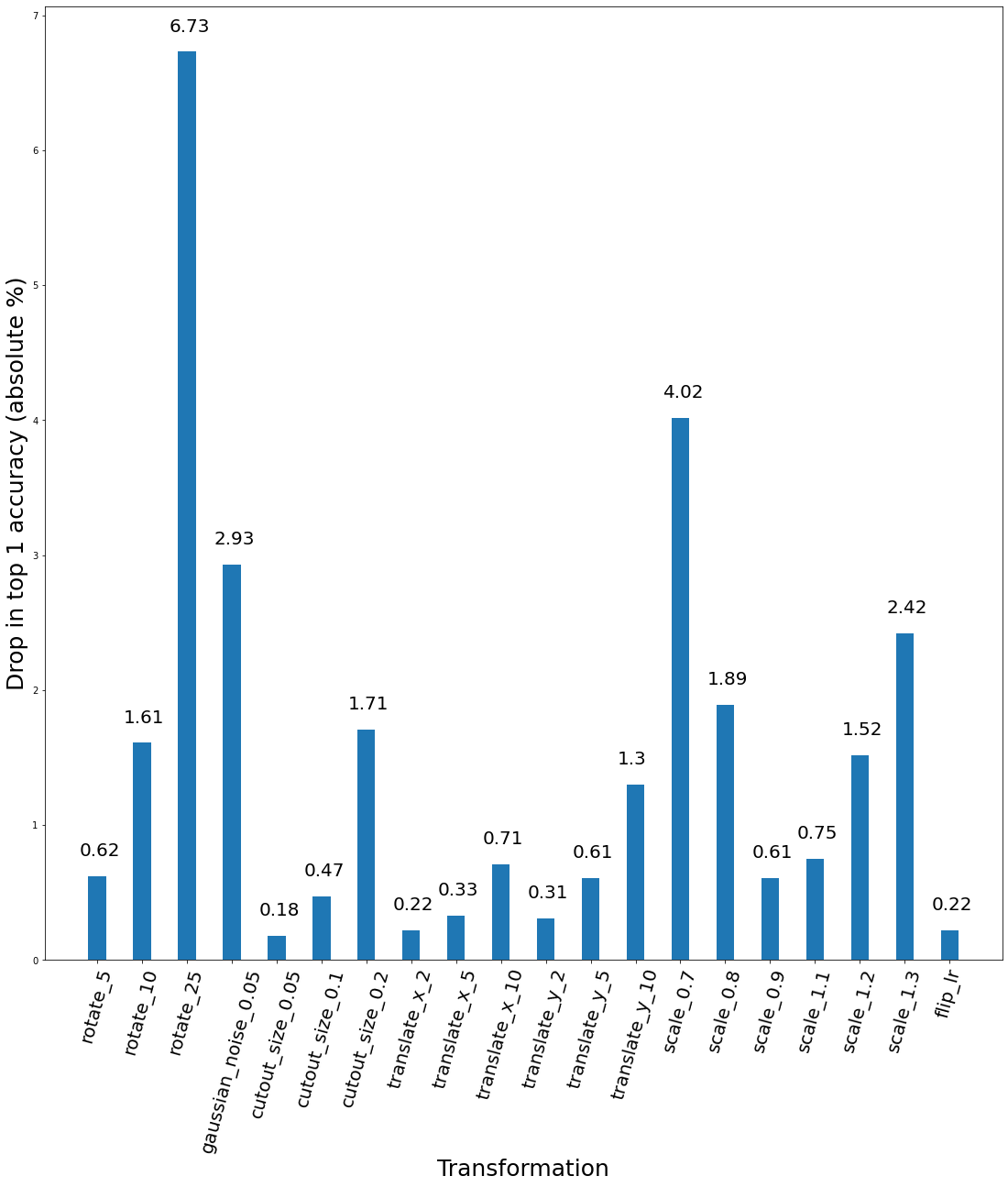}
\caption{Drop in Top-1 accuracy on ImageNet val set for ResNet-50 on various transformations}
\label{fig:results-resnet-50}
\end{figure}

\begin{figure}[!h]
\centering
\includegraphics[width=0.45\textwidth]{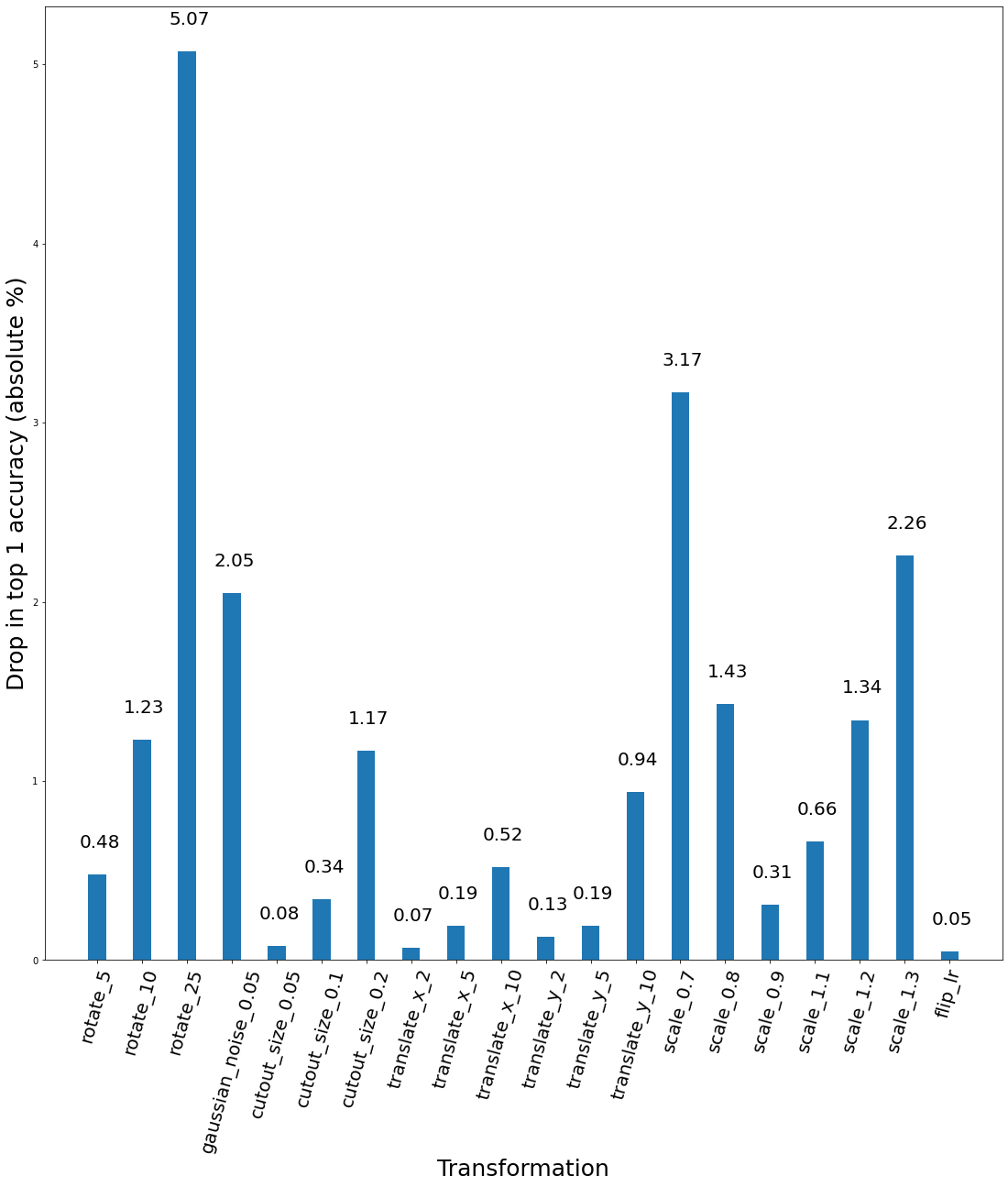}
\caption{Drop in Top-1 accuracy on ImageNet val set for ResNet-152 on various transformations}
\label{fig:results-resnet-152}
\vspace{0.5cm}
\end{figure}

\begin{figure}[!h]
\centering
\includegraphics[width=0.45\textwidth]{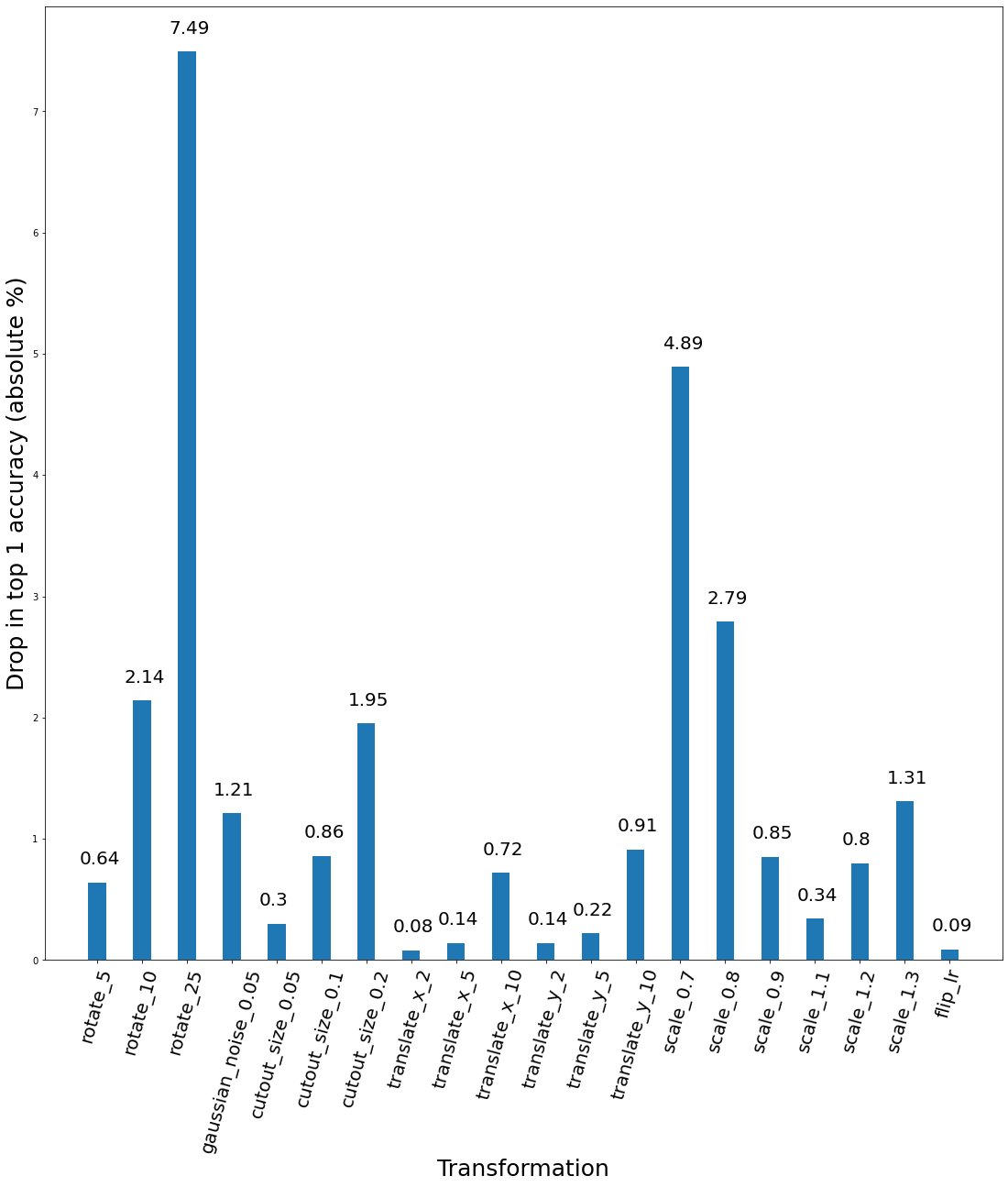}
\caption{Drop in Top-1 accuracy on ImageNet val set for ViT-B/16 on various transformations}
\label{fig:results-vit-b-16}
\end{figure}

\begin{figure}[!h]
\centering
\includegraphics[width=0.45\textwidth]{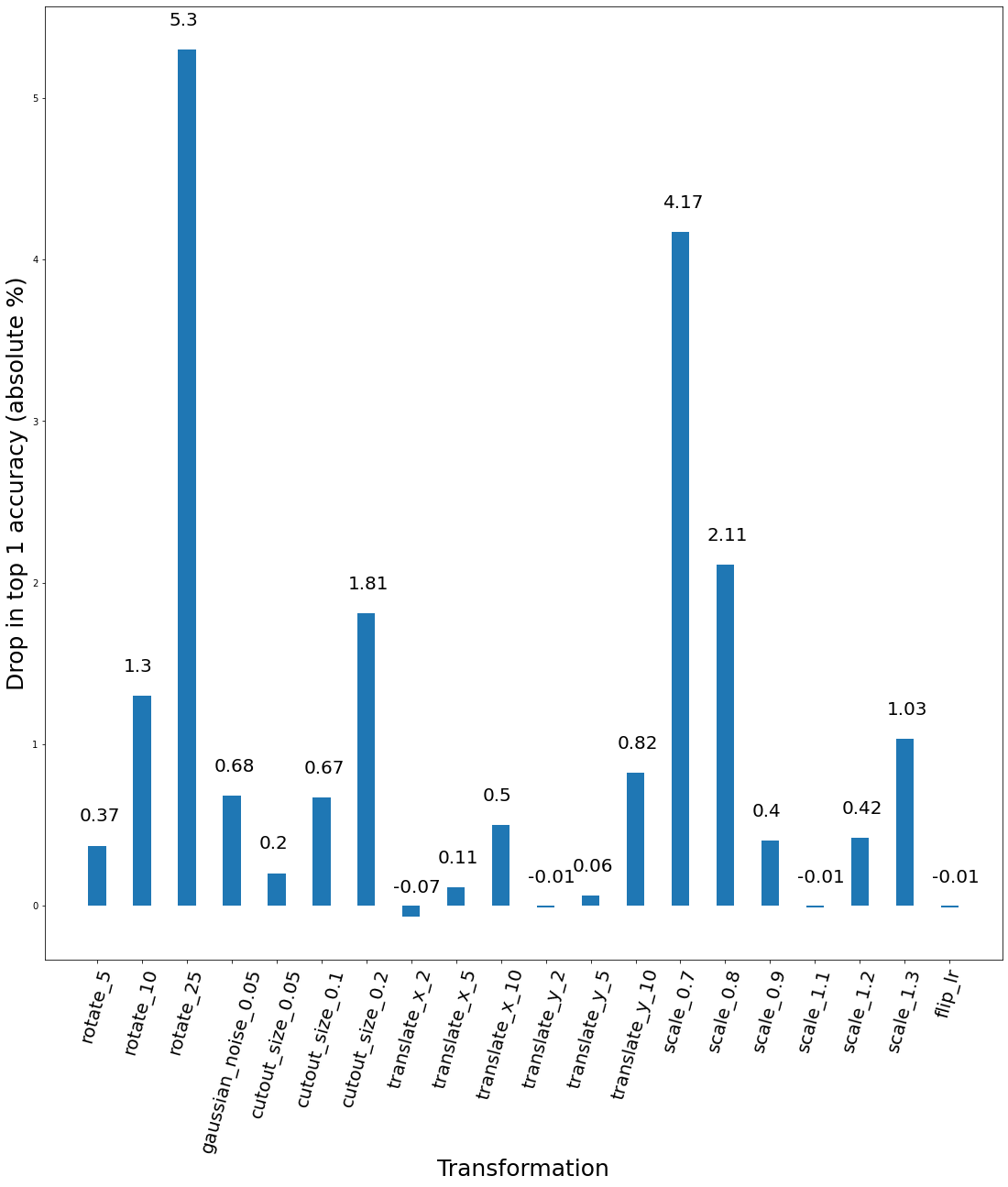}
\caption{Drop in Top-1 accuracy on ImageNet val set for ViT-L/16 on various transformations}
\label{fig:results-vit-l-16}
\vspace{0.5cm}
\end{figure}

We perform various simple transformations to the images in the validation set before normalizing the image and then report the accuracy of the models. The transformations are simple but still degrade the performance significantly. We present the results of a few selected transformations in this section, but the complete set of results are listed down in Appendix \ref{sec:numbers}. We report the drop in top-1 accuracy of ResNet-50 (fig. \ref{fig:results-resnet-50}), ResNet-152 (fig. \ref{fig:results-resnet-152}), ViT-B/16 (fig. \ref{fig:results-vit-b-16}) and ViT-L/16 (fig. \ref{fig:results-vit-l-16}).

\section{Discussion}
\label{sec:discussion}
We found that these simple transformations, which don't change the data distribution significantly, still degrade the performance of the model. We try to bring forward the question of how significant is the increment of a few \% points in accuracies?
We frequently come across papers, which improve the performance over the existing methods by numbers much less than the drop we saw using simple transformations on the validation set. While the paper still could have interesting ideas to be shared with the community, we wonder how much credit should be given to the contribution that the performance increased by a few percentages (or in some case even less than that) over existing methods? Because as it turns out, simple transformations on the images can lead to a change of a few \% in the top-1 accuracy.

\vspace{0.2cm}
\noindent\small{\textbf{Acknowledgements}. We are grateful for the inputs from Meghna Bhatnagar and Nate Knauf. The views and conclusions contained herein are those of the authors and should not be interpreted as necessarily representing the official policies or endorsements, either expressed or implied, of any institute, or any sponsor}
\newpage

{\small
\bibliographystyle{ieee_fullname}
\bibliography{main}

\begin{thebibliography}{1}\itemsep=-1pt

\bibitem{imgaug}
Imgaug, \url{https://imgaug.readthedocs.io/en/latest/}.

\bibitem{papersithcode}
Papers with code - imagenet benchmark (image classification),
  \url{https://paperswithcode.com/sota/image-classification-on-imagenet}.

\bibitem{pytorch}
Pytorch models and pre-trained weights,
  \url{https://pytorch.org/vision/stable/models.html}.

\bibitem{ObjectNet}
Andrei Barbu, David Mayo, Julian Alverio, William Luo, Christopher Wang, Dan
  Gutfreund, Josh Tenenbaum, and Boris Katz.
\newblock Objectnet: A large-scale bias-controlled dataset for pushing the
  limits of object recognition models.
\newblock In H. Wallach, H. Larochelle, A. Beygelzimer, F. d\textquotesingle
  Alch\'{e}-Buc, E. Fox, and R. Garnett, editors, {\em Advances in Neural
  Information Processing Systems}, volume~32. Curran Associates, Inc., 2019.

\bibitem{imagenet-rendition}
Dan Hendrycks, Steven Basart, Norman Mu, Saurav Kadavath, Frank Wang, Evan
  Dorundo, Rahul Desai, Tyler Zhu, Samyak Parajuli, Mike Guo, et~al.
\newblock The many faces of robustness: A critical analysis of
  out-of-distribution generalization.
\newblock In {\em Proceedings of the IEEE/CVF International Conference on
  Computer Vision}, pages 8340--8349, 2021.

\bibitem{imagenet-a}
Dan Hendrycks, Kevin Zhao, Steven Basart, Jacob Steinhardt, and Dawn Song.
\newblock Natural adversarial examples.
\newblock In {\em Proceedings of the IEEE/CVF Conference on Computer Vision and
  Pattern Recognition}, pages 15262--15271, 2021.

\bibitem{generalization-imagenet}
Benjamin Recht, Rebecca Roelofs, Ludwig Schmidt, and Vaishaal Shankar.
\newblock Do {I}mage{N}et classifiers generalize to {I}mage{N}et?
\newblock In Kamalika Chaudhuri and Ruslan Salakhutdinov, editors, {\em
  Proceedings of the 36th International Conference on Machine Learning},
  volume~97 of {\em Proceedings of Machine Learning Research}, pages
  5389--5400. PMLR, 09--15 Jun 2019.

\bibitem{imagenetv2}
Benjamin Recht, Rebecca Roelofs, Ludwig Schmidt, and Vaishaal Shankar.
\newblock Do imagenet classifiers generalize to imagenet?
\newblock In {\em International Conference on Machine Learning}, pages
  5389--5400. PMLR, 2019.

\bibitem{imagenet-sketch}
Haohan Wang, Songwei Ge, Zachary Lipton, and Eric~P Xing.
\newblock Learning robust global representations by penalizing local predictive
  power.
\newblock {\em Advances in Neural Information Processing Systems}, 32, 2019.

\end{thebibliography}
}
\newpage

\appendix
\section{Reproducing numbers}
\label{sec:original}
We first verify that we are able to reproduce the results mentioned on the PyTorch vision website \cite{pytorch}. This verifies that the scaling, loading and other preprocessing is working properly. Comparison is compiled in table \ref{tab:reproduce}.
\begin{table*}[h]
\begin{tabular}{|l|l|l|l|l|l|l|}
\hline
Models    & Top 1 accuracy & Top 1 reported accuracy & \% difference & Top 5 accuracy & Top 5 reported accuracy &\% difference \\ \hline
ResNet-50  & 76.128         & 76.13                                                              & 0.003\%           & 92.862 & 92.862                                                             & 0.000\%           \\ \hline
ResNet-152 & 78.308         & 78.312                                                             & 0.005\%           & 94.046 & 94.046                                                             & 0.000\%           \\ \hline
ViT-B/16 & 76.56         & -                                                              & -           & 93.664 & -                                                             & -          \\ \hline
ViT-L/16 & 78.054         & -                                                              & -           & 94.194 & -                                                             & -          \\ \hline
\end{tabular}
\caption{The top-1 and top-5 accuracies evaluated as compared to the ones reported on the PyTorch vision website \cite{pytorch}.}
\label{tab:reproduce}
\end{table*}

\section{Numbers from all models and transformations}
\label{sec:numbers}
We present all the numbers here for all the models and transformations in tables \ref{tab:resnet50-1} to \ref{tab:vit-l-16-2}.
\begin{table*}
\centering
\begin{tabular}{|l|l|r|r|r|r|}
\hline
Transformation & \begin{tabular}[c]{@{}l@{}}Degree of\\  transformation\end{tabular} & \multicolumn{1}{l|}{Top1} & \multicolumn{1}{l|}{Top5} & \multicolumn{1}{l|}{\begin{tabular}[c]{@{}l@{}}Drop in Top1 \\ accuracy \\ (absolute \%)\end{tabular}} & \multicolumn{1}{l|}{\begin{tabular}[c]{@{}l@{}}Drop in Top5 \\ accuracy \\ (absolute \%)\end{tabular}} \\ \hline
\multirow{9}{*}{Rotation} & 5° & 75.508 & 92.488 & -0.62 & -0.37 \\ \cline{2-6} 
 & 10° & 74.518 & 91.962 & -1.61 & -0.9 \\ \cline{2-6} 
 & 25° & 69.402 & 88.538 & -6.73 & -4.32 \\ \cline{2-6} 
 & 45° & 60.852 & 82.312 & -15.28 & -10.55 \\ \cline{2-6} 
 & 90° & 49.706 & 73.064 & -26.42 & -19.8 \\ \cline{2-6} 
 & -10° & 74.482 & 91.854 & -1.65 & -1.01 \\ \cline{2-6} 
 & -25° & 69.468 & 88.628 & -6.66 & -4.23 \\ \cline{2-6} 
 & -45° & 61.24 & 82.616 & -14.89 & -10.25 \\ \cline{2-6} 
 & -90° & 50.14 & 73.242 & -25.99 & -19.62 \\ \hline
\multirow{7}{*}{Additive Gaussian noise} & variance = 0.05*255 & 73.202 & 91.342 & -2.93 & -1.52 \\ \cline{2-6} 
 & variance = 0.1*255 & 68.23 & 87.988 & -7.9 & -4.87 \\ \cline{2-6} 
 & variance = 0.2*255 & 52.946 & 76.312 & -23.18 & -16.55 \\ \cline{2-6} 
 & variance = 0.3*255 & 35.564 & 58.882 & -40.56 & -33.98 \\ \cline{2-6} 
 & variance = 0.4*255 & 21.896 & 40.99 & -54.23 & -51.87 \\ \cline{2-6} 
 & variance = 0.5*255 & 12.866 & 26.87 & -63.26 & -65.99 \\ \cline{2-6} 
 & variance = 1*255 & 1.436 & 4.086 & -74.69 & -88.78 \\ \hline
\multirow{6}{*}{Rectangular Cutouts} & size = 0.05*(H, W) & 75.95 & 92.786 & -0.18 & -0.08 \\ \cline{2-6} 
 & size = 0.1*(H, W) & 75.658 & 92.622 & -0.47 & -0.24 \\ \cline{2-6} 
 & size = 0.2*(H, W) & 74.422 & 91.896 & -1.71 & -0.97 \\ \cline{2-6} 
 & size = 0.3*(H, W) & 72.848 & 90.856 & -3.28 & -2.01 \\ \cline{2-6} 
 & size = 0.4*(H, W) & 70.278 & 89.168 & -5.85 & -3.69 \\ \cline{2-6} 
 & size = 0.5*(H, W) & 66.43 & 86.296 & -9.7 & -6.57 \\ \hline
\multirow{8}{*}{\begin{tabular}[c]{@{}l@{}}Horizontal translation\\ (to the right)\end{tabular}} & 0.02*W & 75.904 & 92.822 & -0.22 & -0.04 \\ \cline{2-6} 
 & 0.05*W & 75.798 & 92.704 & -0.33 & -0.16 \\ \cline{2-6} 
 & 0.1*W & 75.414 & 92.392 & -0.71 & -0.47 \\ \cline{2-6} 
 & 0.15*W & 74.462 & 91.85 & -1.67 & -1.01 \\ \cline{2-6} 
 & 0.2*W & 73.006 & 90.954 & -3.12 & -1.91 \\ \cline{2-6} 
 & 0.25*W & 71.174 & 89.586 & -4.95 & -3.28 \\ \cline{2-6} 
 & 0.35*W & 64.514 & 84.99 & -11.61 & -7.87 \\ \cline{2-6} 
 & 0.5*W & 46.83 & 68.95 & -29.3 & -23.91 \\ \hline
\multirow{8}{*}{\begin{tabular}[c]{@{}l@{}}Horizontal translation\\ (to the left)\end{tabular}} & 0.02*W & 75.918 & 92.852 & -0.21 & -0.01 \\ \cline{2-6} 
 & 0.05*W & 75.856 & 92.806 & -0.27 & -0.06 \\ \cline{2-6} 
 & 0.1*W & 75.328 & 92.502 & -0.8 & -0.36 \\ \cline{2-6} 
 & 0.15*W & 74.54 & 91.922 & -1.59 & -0.94 \\ \cline{2-6} 
 & 0.2*W & 73.13 & 91.104 & -3 & -1.76 \\ \cline{2-6} 
 & 0.25*W & 71.25 & 89.87 & -4.88 & -2.99 \\ \cline{2-6} 
 & 0.35*W & 65.052 & 85.258 & -11.08 & -7.6 \\ \cline{2-6} 
 & 0.5*W & 48.048 & 70.06 & -28.08 & -22.8 \\ \hline
\end{tabular}
\caption{Results for all the transformations for ResNet-50 (part 1 of 2)}
\label{tab:resnet50-1}
\end{table*}

\begin{table*}
\centering
\begin{tabular}{|l|l|r|r|r|r|}
\hline
Transformation & \begin{tabular}[c]{@{}l@{}}Degree of\\  transformation\end{tabular} & \multicolumn{1}{l|}{Top1} & \multicolumn{1}{l|}{Top5} & \multicolumn{1}{l|}{\begin{tabular}[c]{@{}l@{}}Drop in Top1 \\ accuracy \\ (absolute \%)\end{tabular}} & \multicolumn{1}{l|}{\begin{tabular}[c]{@{}l@{}}Drop in Top5 \\ accuracy \\ (absolute \%)\end{tabular}} \\ \hline
\multirow{8}{*}{\begin{tabular}[c]{@{}l@{}}Vertical translation\\ (upwards)\end{tabular}} & 0.02*H & 75.82 & 92.772 & -0.31 & -0.09 \\ \cline{2-6} 
 & 0.05*H & 75.514 & 92.734 & -0.61 & -0.13 \\ \cline{2-6} 
 & 0.1*H & 74.828 & 92.176 & -1.3 & -0.69 \\ \cline{2-6} 
 & 0.15*H & 73.936 & 91.652 & -2.19 & -1.21 \\ \cline{2-6} 
 & 0.2*H & 72.352 & 90.628 & -3.78 & -2.23 \\ \cline{2-6} 
 & 0.25*H & 70.374 & 89.318 & -5.75 & -3.54 \\ \cline{2-6} 
 & 0.35*H & 64.48 & 84.496 & -11.65 & -8.37 \\ \cline{2-6} 
 & 0.5*H & 48.052 & 70.054 & -28.08 & -22.81 \\ \hline
\multirow{8}{*}{\begin{tabular}[c]{@{}l@{}}Vertical translation\\ (downwards)\end{tabular}} & 0.02*H & 75.75 & 92.818 & -0.38 & -0.04 \\ \cline{2-6} 
 & 0.05*H & 75.71 & 92.692 & -0.42 & -0.17 \\ \cline{2-6} 
 & 0.1*H & 74.874 & 92.352 & -1.25 & -0.51 \\ \cline{2-6} 
 & 0.15*H & 74.072 & 91.762 & -2.06 & -1.1 \\ \cline{2-6} 
 & 0.2*H & 72.632 & 90.896 & -3.5 & -1.97 \\ \cline{2-6} 
 & 0.25*H & 70.476 & 89.54 & -5.65 & -3.32 \\ \cline{2-6} 
 & 0.35*H & 64.534 & 85.448 & -11.59 & -7.41 \\ \cline{2-6} 
 & 0.5*H & 48.628 & 72.396 & -27.5 & -20.47 \\ \hline
\multirow{15}{*}{\begin{tabular}[c]{@{}l@{}}Scale\\ (Zoom-in/Zoom-out)\end{tabular}} & 0.2*(H, W) & 4.744 & 11.05 & -71.38 & -81.81 \\ \cline{2-6} 
 & 0.3*(H, W) & 25.064 & 45.194 & -51.06 & -47.67 \\ \cline{2-6} 
 & 0.4*(H, W) & 45.668 & 69.58 & -30.46 & -23.28 \\ \cline{2-6} 
 & 0.5*(H, W) & 59.5 & 82.108 & -16.63 & -10.75 \\ \cline{2-6} 
 & 0.6*(H, W) & 67.484 & 87.654 & -8.64 & -5.21 \\ \cline{2-6} 
 & 0.7*(H, W) & 72.112 & 90.564 & -4.02 & -2.3 \\ \cline{2-6} 
 & 0.8*(H, W) & 74.234 & 91.972 & -1.89 & -0.89 \\ \cline{2-6} 
 & 0.9*(H, W) & 75.516 & 92.674 & -0.61 & -0.19 \\ \cline{2-6} 
 & 1.1*(H, W) & 75.378 & 92.384 & -0.75 & -0.48 \\ \cline{2-6} 
 & 1.2*(H, W) & 74.604 & 91.942 & -1.52 & -0.92 \\ \cline{2-6} 
 & 1.3*(H, W) & 73.712 & 91.218 & -2.42 & -1.64 \\ \cline{2-6} 
 & 1.4*(H, W) & 72.662 & 90.404 & -3.47 & -2.46 \\ \cline{2-6} 
 & 1.5*(H, W) & 71.302 & 89.748 & -4.83 & -3.11 \\ \cline{2-6} 
 & 1.6*(H, W) & 70.036 & 88.834 & -6.09 & -4.03 \\ \cline{2-6} 
 & 1.75*(H, W) & 68.094 & 87.568 & -8.03 & -5.29 \\ \hline
Flip left-right & - & 75.906 & 92.872 & -0.22 & 0.01 \\ \hline
Flip up-down & - & 51.566 & 75.886 & -24.56 & -16.98 \\ \hline
\end{tabular}
\caption{Results for all the transformations for ResNet-50 (part 2 of 2)}
\label{tab:resnet50-2}
\end{table*}
\begin{table*}
\centering
\begin{tabular}{|l|l|r|r|r|r|}
\hline
Transformation & \begin{tabular}[c]{@{}l@{}}Degree of\\  transformation\end{tabular} & \multicolumn{1}{l|}{Top1} & \multicolumn{1}{l|}{Top5} & \multicolumn{1}{l|}{\begin{tabular}[c]{@{}l@{}}Drop in Top1 \\ accuracy \\ (absolute \%)\end{tabular}} & \multicolumn{1}{l|}{\begin{tabular}[c]{@{}l@{}}Drop in Top5 \\ accuracy \\ (absolute \%)\end{tabular}} \\ \hline
\multirow{9}{*}{Rotation} & 5° & 77.824 & 93.86 & -0.48 & -0.19 \\ \cline{2-6} 
 & 10° & 77.082 & 93.374 & -1.23 & -0.67 \\ \cline{2-6} 
 & 25° & 73.236 & 91.274 & -5.07 & -2.77 \\ \cline{2-6} 
 & 45° & 65.898 & 86.41 & -12.41 & -7.64 \\ \cline{2-6} 
 & 90° & 53.388 & 75.95 & -24.92 & -18.1 \\ \cline{2-6} 
 & -10° & 76.932 & 93.382 & -1.38 & -0.66 \\ \cline{2-6} 
 & -25° & 73.134 & 91.146 & -5.17 & -2.9 \\ \cline{2-6} 
 & -45° & 65.682 & 86.268 & -12.63 & -7.78 \\ \cline{2-6} 
 & -90° & 53.44 & 76.21 & -24.87 & -17.84 \\ \hline
\multirow{7}{*}{Additive Gaussian noise} & variance = 0.05*255 & 76.254 & 93.054 & -2.05 & -0.99 \\ \cline{2-6} 
 & variance = 0.1*255 & 72.326 & 90.87 & -5.98 & -3.18 \\ \cline{2-6} 
 & variance = 0.2*255 & 61.596 & 83.442 & -16.71 & -10.6 \\ \cline{2-6} 
 & variance = 0.3*255 & 47.246 & 70.624 & -31.06 & -23.42 \\ \cline{2-6} 
 & variance = 0.4*255 & 33.282 & 55.344 & -45.03 & -38.7 \\ \cline{2-6} 
 & variance = 0.5*255 & 22.166 & 40.488 & -56.14 & -53.56 \\ \cline{2-6} 
 & variance = 1*255 & 3.086 & 7.332 & -75.22 & -86.71 \\ \hline
\multirow{6}{*}{Rectangular Cutouts} & size = 0.05*(H, W) & 78.226 & 93.948 & -0.08 & -0.1 \\ \cline{2-6} 
 & size = 0.1*(H, W) & 77.968 & 93.856 & -0.34 & -0.19 \\ \cline{2-6} 
 & size = 0.2*(H, W) & 77.134 & 93.408 & -1.17 & -0.64 \\ \cline{2-6} 
 & size = 0.3*(H, W) & 75.542 & 92.616 & -2.77 & -1.43 \\ \cline{2-6} 
 & size = 0.4*(H, W) & 73.41 & 91.212 & -4.9 & -2.83 \\ \cline{2-6} 
 & size = 0.5*(H, W) & 69.984 & 88.39 & -8.32 & -5.66 \\ \hline
\multirow{8}{*}{\begin{tabular}[c]{@{}l@{}}Horizontal translation\\ (to the right)\end{tabular}} & 0.02*W & 78.242 & 94.05 & -0.07 & 0 \\ \cline{2-6} 
 & 0.05*W & 78.12 & 93.998 & -0.19 & -0.05 \\ \cline{2-6} 
 & 0.1*W & 77.788 & 93.716 & -0.52 & -0.33 \\ \cline{2-6} 
 & 0.15*W & 77 & 93.274 & -1.31 & -0.77 \\ \cline{2-6} 
 & 0.2*W & 75.79 & 92.57 & -2.52 & -1.48 \\ \cline{2-6} 
 & 0.25*W & 74.002 & 91.466 & -4.31 & -2.58 \\ \cline{2-6} 
 & 0.35*W & 67.772 & 87.1 & -10.54 & -6.95 \\ \cline{2-6} 
 & 0.5*W & 50.178 & 72.004 & -28.13 & -22.04 \\ \hline
\multirow{8}{*}{\begin{tabular}[c]{@{}l@{}}Horizontal translation\\ (to the left)\end{tabular}} & 0.02*W & 78.276 & 93.984 & -0.03 & -0.06 \\ \cline{2-6} 
 & 0.05*W & 78.194 & 93.846 & -0.11 & -0.2 \\ \cline{2-6} 
 & 0.1*W & 77.604 & 93.636 & -0.7 & -0.41 \\ \cline{2-6} 
 & 0.15*W & 76.914 & 93.338 & -1.39 & -0.71 \\ \cline{2-6} 
 & 0.2*W & 75.824 & 92.598 & -2.48 & -1.45 \\ \cline{2-6} 
 & 0.25*W & 74.138 & 91.506 & -4.17 & -2.54 \\ \cline{2-6} 
 & 0.35*W & 68.134 & 87.318 & -10.17 & -6.73 \\ \cline{2-6} 
 & 0.5*W & 51.284 & 72.872 & -27.02 & -21.17 \\ \hline
\end{tabular}
\caption{Results for all the transformations for ResNet-152 (part 1 of 2)}
\label{tab:resnet152-1}
\end{table*}

\begin{table*}
\centering
\begin{tabular}{|l|l|r|r|r|r|}
\hline
Transformation & \begin{tabular}[c]{@{}l@{}}Degree of\\  transformation\end{tabular} & \multicolumn{1}{l|}{Top1} & \multicolumn{1}{l|}{Top5} & \multicolumn{1}{l|}{\begin{tabular}[c]{@{}l@{}}Drop in Top1 \\ accuracy \\ (absolute \%)\end{tabular}} & \multicolumn{1}{l|}{\begin{tabular}[c]{@{}l@{}}Drop in Top5 \\ accuracy \\ (absolute \%)\end{tabular}} \\ \hline
\multirow{8}{*}{\begin{tabular}[c]{@{}l@{}}Vertical translation\\ (upwards)\end{tabular}} & 0.02*H & 78.174 & 94.006 & -0.13 & -0.04 \\ \cline{2-6} 
 & 0.05*H & 78.116 & 93.9 & -0.19 & -0.15 \\ \cline{2-6} 
 & 0.1*H & 77.366 & 93.598 & -0.94 & -0.45 \\ \cline{2-6} 
 & 0.15*H & 76.466 & 92.894 & -1.84 & -1.15 \\ \cline{2-6} 
 & 0.2*H & 75.296 & 92.192 & -3.01 & -1.85 \\ \cline{2-6} 
 & 0.25*H & 73.288 & 90.818 & -5.02 & -3.23 \\ \cline{2-6} 
 & 0.35*H & 67.662 & 86.832 & -10.65 & -7.21 \\ \cline{2-6} 
 & 0.5*H & 51.534 & 72.856 & -26.77 & -21.19 \\ \hline
\multirow{8}{*}{\begin{tabular}[c]{@{}l@{}}Vertical translation\\ (downwards)\end{tabular}} & 0.02*H & 78.254 & 94.004 & -0.05 & -0.04 \\ \cline{2-6} 
 & 0.05*H & 78.128 & 93.986 & -0.18 & -0.06 \\ \cline{2-6} 
 & 0.1*H & 77.506 & 93.612 & -0.8 & -0.43 \\ \cline{2-6} 
 & 0.15*H & 76.608 & 93.18 & -1.7 & -0.87 \\ \cline{2-6} 
 & 0.2*H & 75.454 & 92.348 & -2.85 & -1.7 \\ \cline{2-6} 
 & 0.25*H & 73.428 & 91.256 & -4.88 & -2.79 \\ \cline{2-6} 
 & 0.35*H & 67.55 & 87.522 & -10.76 & -6.52 \\ \cline{2-6} 
 & 0.5*H & 52.126 & 75.448 & -26.18 & -18.6 \\ \hline
\multirow{15}{*}{\begin{tabular}[c]{@{}l@{}}Scale\\ (Zoom-in/Zoom-out)\end{tabular}} & 0.2*(H, W) & 7.53 & 17.132 & -70.78 & -76.91 \\ \cline{2-6} 
 & 0.3*(H, W) & 31.306 & 53.752 & -47 & -40.29 \\ \cline{2-6} 
 & 0.4*(H, W) & 52.4 & 75.946 & -25.91 & -18.1 \\ \cline{2-6} 
 & 0.5*(H, W) & 64.454 & 85.672 & -13.85 & -8.37 \\ \cline{2-6} 
 & 0.6*(H, W) & 71.352 & 90.18 & -6.96 & -3.87 \\ \cline{2-6} 
 & 0.7*(H, W) & 75.142 & 92.594 & -3.17 & -1.45 \\ \cline{2-6} 
 & 0.8*(H, W) & 76.88 & 93.53 & -1.43 & -0.52 \\ \cline{2-6} 
 & 0.9*(H, W) & 77.996 & 93.972 & -0.31 & -0.07 \\ \cline{2-6} 
 & 1.1*(H, W) & 77.648 & 93.656 & -0.66 & -0.39 \\ \cline{2-6} 
 & 1.2*(H, W) & 76.97 & 93.256 & -1.34 & -0.79 \\ \cline{2-6} 
 & 1.3*(H, W) & 76.048 & 92.758 & -2.26 & -1.29 \\ \cline{2-6} 
 & 1.4*(H, W) & 74.996 & 92.076 & -3.31 & -1.97 \\ \cline{2-6} 
 & 1.5*(H, W) & 74.056 & 91.442 & -4.25 & -2.6 \\ \cline{2-6} 
 & 1.6*(H, W) & 72.856 & 90.598 & -5.45 & -3.45 \\ \cline{2-6} 
 & 1.75*(H, W) & 71.25 & 89.414 & -7.06 & -4.63 \\ \hline
Flip left-right & - & 78.254 & 94.026 & -0.05 & -0.02 \\ \hline
Flip up-down & - & 54.086 & 78.06 & -24.22 & -15.99 \\ \hline
\end{tabular}
\caption{Results for all the transformations for ResNet-152  (part 2 of 2)}
\label{tab:resnet152-2}
\end{table*}
\begin{table*}
\centering
\begin{tabular}{|l|l|r|r|r|r|}
\hline
Transformation & \begin{tabular}[c]{@{}l@{}}Degree of\\  transformation\end{tabular} & \multicolumn{1}{l|}{Top1} & \multicolumn{1}{l|}{Top5} & \multicolumn{1}{l|}{\begin{tabular}[c]{@{}l@{}}Drop in Top1 \\ accuracy \\ (absolute \%)\end{tabular}} & \multicolumn{1}{l|}{\begin{tabular}[c]{@{}l@{}}Drop in Top5 \\ accuracy \\ (absolute \%)\end{tabular}} \\ \hline
\multirow{9}{*}{Rotation} & 5° & 75.924 & 93.276 & -0.64 & -0.39 \\ \cline{2-6} 
 & 10° & 74.428 & 92.34 & -2.14 & -1.32 \\ \cline{2-6} 
 & 25° & 69.078 & 88.904 & -7.49 & -4.76 \\ \cline{2-6} 
 & 45° & 59.888 & 81.878 & -16.68 & -11.79 \\ \cline{2-6} 
 & 90° & 46.526 & 69.162 & -30.04 & -24.5 \\ \cline{2-6} 
 & -10° & 74.386 & 92.262 & -2.18 & -1.4 \\ \cline{2-6} 
 & -25° & 69.188 & 89.05 & -7.38 & -4.61 \\ \cline{2-6} 
 & -45° & 59.81 & 81.634 & -16.76 & -12.03 \\ \cline{2-6} 
 & -90° & 46.76 & 68.94 & -29.81 & -24.72 \\ \hline
\multirow{7}{*}{Additive Gaussian noise} & variance = 0.05*255 & 75.36 & 92.942 & -1.21 & -0.72 \\ \cline{2-6} 
 & variance = 0.1*255 & 71.946 & 90.862 & -4.62 & -2.8 \\ \cline{2-6} 
 & variance = 0.2*255 & 62.946 & 84.394 & -13.62 & -9.27 \\ \cline{2-6} 
 & variance = 0.3*255 & 53.122 & 75.564 & -23.44 & -18.1 \\ \cline{2-6} 
 & variance = 0.4*255 & 42.58 & 64.982 & -33.99 & -28.68 \\ \cline{2-6} 
 & variance = 0.5*255 & 32.398 & 53.074 & -44.17 & -40.59 \\ \cline{2-6} 
 & variance = 1*255 & 5.652 & 12.354 & -70.91 & -81.31 \\ \hline
\multirow{6}{*}{Rectangular Cutouts} & size = 0.05*(H, W) & 76.27 & 93.496 & -0.3 & -0.17 \\ \cline{2-6} 
 & size = 0.1*(H, W) & 75.708 & 93.32 & -0.86 & -0.34 \\ \cline{2-6} 
 & size = 0.2*(H, W) & 74.612 & 92.576 & -1.95 & -1.09 \\ \cline{2-6} 
 & size = 0.3*(H, W) & 73.064 & 91.498 & -3.5 & -2.17 \\ \cline{2-6} 
 & size = 0.4*(H, W) & 70.62 & 89.568 & -5.95 & -4.1 \\ \cline{2-6} 
 & size = 0.5*(H, W) & 67.14 & 86.854 & -9.43 & -6.81 \\ \hline
\multirow{8}{*}{\begin{tabular}[c]{@{}l@{}}Horizontal translation\\ (to the right)\end{tabular}} & 0.02*W & 76.49 & 93.62 & -0.08 & -0.04 \\ \cline{2-6} 
 & 0.05*W & 76.428 & 93.514 & -0.14 & -0.15 \\ \cline{2-6} 
 & 0.1*W & 75.85 & 93.29 & -0.72 & -0.37 \\ \cline{2-6} 
 & 0.15*W & 75.192 & 92.74 & -1.37 & -0.92 \\ \cline{2-6} 
 & 0.2*W & 73.914 & 91.786 & -2.65 & -1.88 \\ \cline{2-6} 
 & 0.25*W & 71.974 & 90.44 & -4.59 & -3.22 \\ \cline{2-6} 
 & 0.35*W & 65.61 & 85.934 & -10.96 & -7.73 \\ \cline{2-6} 
 & 0.5*W & 48.49 & 70.268 & -28.08 & -23.4 \\ \hline
\multirow{8}{*}{\begin{tabular}[c]{@{}l@{}}Horizontal translation\\ (to the left)\end{tabular}} & 0.02*W & 76.474 & 93.55 & -0.09 & -0.11 \\ \cline{2-6} 
 & 0.05*W & 76.392 & 93.526 & -0.17 & -0.14 \\ \cline{2-6} 
 & 0.1*W & 75.696 & 93.088 & -0.87 & -0.58 \\ \cline{2-6} 
 & 0.15*W & 75.004 & 92.526 & -1.56 & -1.14 \\ \cline{2-6} 
 & 0.2*W & 73.558 & 91.624 & -3.01 & -2.04 \\ \cline{2-6} 
 & 0.25*W & 71.71 & 90.542 & -4.86 & -3.12 \\ \cline{2-6} 
 & 0.35*W & 65.472 & 85.522 & -11.09 & -8.14 \\ \cline{2-6} 
 & 0.5*W & 48.484 & 70.702 & -28.08 & -22.96 \\ \hline
\end{tabular}
\caption{Results for all the transformations for ViT-B/16 (part 1 of 2)}
\label{tab:vit-b-16-1}
\end{table*}

\begin{table*}
\centering
\begin{tabular}{|l|l|r|r|r|r|}
\hline
Transformation & \begin{tabular}[c]{@{}l@{}}Degree of\\  transformation\end{tabular} & \multicolumn{1}{l|}{Top1} & \multicolumn{1}{l|}{Top5} & \multicolumn{1}{l|}{\begin{tabular}[c]{@{}l@{}}Drop in Top1 \\ accuracy \\ (absolute \%)\end{tabular}} & \multicolumn{1}{l|}{\begin{tabular}[c]{@{}l@{}}Drop in Top5 \\ accuracy \\ (absolute \%)\end{tabular}} \\ \hline
\multirow{8}{*}{\begin{tabular}[c]{@{}l@{}}Vertical translation\\ (upwards)\end{tabular}} & 0.02*H & 76.426 & 93.522 & -0.14 & -0.14 \\ \cline{2-6} 
 & 0.05*H & 76.346 & 93.318 & -0.22 & -0.35 \\ \cline{2-6} 
 & 0.1*H & 75.66 & 93.032 & -0.91 & -0.63 \\ \cline{2-6} 
 & 0.15*H & 74.664 & 92.63 & -1.9 & -1.03 \\ \cline{2-6} 
 & 0.2*H & 73.408 & 91.552 & -3.16 & -2.11 \\ \cline{2-6} 
 & 0.25*H & 71.42 & 90.212 & -5.15 & -3.45 \\ \cline{2-6} 
 & 0.35*H & 66.068 & 85.996 & -10.5 & -7.67 \\ \cline{2-6} 
 & 0.5*H & 50.67 & 72.666 & -25.9 & -21 \\ \hline
\multirow{8}{*}{\begin{tabular}[c]{@{}l@{}}Vertical translation\\ (downwards)\end{tabular}} & 0.02*H & 76.456 & 93.478 & -0.11 & -0.19 \\ \cline{2-6} 
 & 0.05*H & 76.366 & 93.546 & -0.2 & -0.12 \\ \cline{2-6} 
 & 0.1*H & 75.44 & 93.016 & -1.13 & -0.65 \\ \cline{2-6} 
 & 0.15*H & 74.56 & 92.488 & -2.01 & -1.18 \\ \cline{2-6} 
 & 0.2*H & 72.946 & 91.57 & -3.62 & -2.09 \\ \cline{2-6} 
 & 0.25*H & 70.984 & 90.496 & -5.58 & -3.17 \\ \cline{2-6} 
 & 0.35*H & 64.838 & 86.16 & -11.73 & -7.5 \\ \cline{2-6} 
 & 0.5*H & 49.36 & 73.18 & -27.21 & -20.48 \\ \hline
\multirow{15}{*}{\begin{tabular}[c]{@{}l@{}}Scale\\ (Zoom-in/Zoom-out)\end{tabular}} & 0.2*(H, W) & 7.086 & 15.544 & -69.48 & -78.12 \\ \cline{2-6} 
 & 0.3*(H, W) & 27.286 & 47.336 & -49.28 & -46.33 \\ \cline{2-6} 
 & 0.4*(H, W) & 46.88 & 69.696 & -29.69 & -23.97 \\ \cline{2-6} 
 & 0.5*(H, W) & 58.824 & 81.006 & -17.74 & -12.66 \\ \cline{2-6} 
 & 0.6*(H, W) & 66.72 & 87.138 & -9.85 & -6.53 \\ \cline{2-6} 
 & 0.7*(H, W) & 71.68 & 90.65 & -4.89 & -3.01 \\ \cline{2-6} 
 & 0.8*(H, W) & 73.774 & 92.166 & -2.79 & -1.5 \\ \cline{2-6} 
 & 0.9*(H, W) & 75.72 & 93.22 & -0.85 & -0.44 \\ \cline{2-6} 
 & 1.1*(H, W) & 76.23 & 93.466 & -0.34 & -0.2 \\ \cline{2-6} 
 & 1.2*(H, W) & 75.764 & 93.234 & -0.8 & -0.43 \\ \cline{2-6} 
 & 1.3*(H, W) & 75.256 & 92.692 & -1.31 & -0.97 \\ \cline{2-6} 
 & 1.4*(H, W) & 74.488 & 92.268 & -2.08 & -1.4 \\ \cline{2-6} 
 & 1.5*(H, W) & 73.634 & 91.63 & -2.93 & -2.03 \\ \cline{2-6} 
 & 1.6*(H, W) & 72.59 & 90.848 & -3.98 & -2.82 \\ \cline{2-6} 
 & 1.75*(H, W) & 71.27 & 89.774 & -5.3 & -3.89 \\ \hline
Flip left-right & - & 76.478 & 93.63 & -0.09 & -0.03 \\ \hline
Flip up-down & - & 46.942 & 70.22 & -29.62 & -23.44 \\ \hline
\end{tabular}
\caption{Results for all the transformations for ViT-B/16 (part 2 of 2)}
\label{tab:vit-b-16-2}
\end{table*}
\begin{table*}
\centering
\begin{tabular}{|l|l|r|r|r|r|}
\hline
Transformation & \begin{tabular}[c]{@{}l@{}}Degree of\\  transformation\end{tabular} & \multicolumn{1}{l|}{Top1} & \multicolumn{1}{l|}{Top5} & \multicolumn{1}{l|}{\begin{tabular}[c]{@{}l@{}}Drop in Top1 \\ accuracy \\ (absolute \%)\end{tabular}} & \multicolumn{1}{l|}{\begin{tabular}[c]{@{}l@{}}Drop in Top5 \\ accuracy \\ (absolute \%)\end{tabular}} \\ \hline
\multirow{9}{*}{Rotation} & 5° & 77.682 & 94.084 & -0.37 & -0.11 \\ \cline{2-6} 
 & 10° & 76.752 & 93.364 & -1.3 & -0.83 \\ \cline{2-6} 
 & 25° & 72.754 & 91.04 & -5.3 & -3.15 \\ \cline{2-6} 
 & 45° & 64.834 & 85.432 & -13.22 & -8.76 \\ \cline{2-6} 
 & 90° & 51.492 & 73.178 & -26.56 & -21.02 \\ \cline{2-6} 
 & -10° & 76.536 & 93.476 & -1.52 & -0.72 \\ \cline{2-6} 
 & -25° & 72.672 & 90.916 & -5.38 & -3.28 \\ \cline{2-6} 
 & -45° & 64.5 & 85.17 & -13.55 & -9.02 \\ \cline{2-6} 
 & -90° & 51.678 & 73.34 & -26.38 & -20.85 \\ \hline
\multirow{7}{*}{Additive Gaussian noise} & variance = 0.05*255 & 77.374 & 93.892 & -0.68 & -0.3 \\ \cline{2-6} 
 & variance = 0.1*255 & 74.468 & 92.162 & -3.59 & -2.03 \\ \cline{2-6} 
 & variance = 0.2*255 & 67.142 & 87.184 & -10.91 & -7.01 \\ \cline{2-6} 
 & variance = 0.3*255 & 58.89 & 80.616 & -19.16 & -13.58 \\ \cline{2-6} 
 & variance = 0.4*255 & 49.834 & 72.162 & -28.22 & -22.03 \\ \cline{2-6} 
 & variance = 0.5*255 & 41.016 & 63.006 & -37.04 & -31.19 \\ \cline{2-6} 
 & variance = 1*255 & 10.462 & 21.236 & -67.59 & -72.96 \\ \hline
\multirow{6}{*}{Rectangular Cutouts} & size = 0.05*(H, W) & 77.852 & 94.086 & -0.2 & -0.11 \\ \cline{2-6} 
 & size = 0.1*(H, W) & 77.386 & 93.832 & -0.67 & -0.36 \\ \cline{2-6} 
 & size = 0.2*(H, W) & 76.248 & 93.24 & -1.81 & -0.95 \\ \cline{2-6} 
 & size = 0.3*(H, W) & 74.648 & 92.176 & -3.41 & -2.02 \\ \cline{2-6} 
 & size = 0.4*(H, W) & 72.24 & 90.45 & -5.81 & -3.74 \\ \cline{2-6} 
 & size = 0.5*(H, W) & 68.588 & 87.61 & -9.47 & -6.58 \\ \hline
\multirow{8}{*}{\begin{tabular}[c]{@{}l@{}}Horizontal translation\\ (to the right)\end{tabular}} & 0.02*W & 78.122 & 94.292 & 0.07 & 0.1 \\ \cline{2-6} 
 & 0.05*W & 77.944 & 94.18 & -0.11 & -0.01 \\ \cline{2-6} 
 & 0.1*W & 77.554 & 93.876 & -0.5 & -0.32 \\ \cline{2-6} 
 & 0.15*W & 76.858 & 93.408 & -1.2 & -0.79 \\ \cline{2-6} 
 & 0.2*W & 75.622 & 92.684 & -2.43 & -1.51 \\ \cline{2-6} 
 & 0.25*W & 73.634 & 91.394 & -4.42 & -2.8 \\ \cline{2-6} 
 & 0.35*W & 68.242 & 87.412 & -9.81 & -6.78 \\ \cline{2-6} 
 & 0.5*W & 50.662 & 71.894 & -27.39 & -22.3 \\ \hline
\multirow{8}{*}{\begin{tabular}[c]{@{}l@{}}Horizontal translation\\ (to the left)\end{tabular}} & 0.02*W & 78.184 & 94.266 & 0.13 & 0.07 \\ \cline{2-6} 
 & 0.05*W & 78.016 & 94.278 & -0.04 & 0.08 \\ \cline{2-6} 
 & 0.1*W & 77.804 & 94.044 & -0.25 & -0.15 \\ \cline{2-6} 
 & 0.15*W & 76.84 & 93.48 & -1.21 & -0.71 \\ \cline{2-6} 
 & 0.2*W & 75.606 & 92.72 & -2.45 & -1.47 \\ \cline{2-6} 
 & 0.25*W & 73.726 & 91.546 & -4.33 & -2.65 \\ \cline{2-6} 
 & 0.35*W & 67.772 & 86.956 & -10.28 & -7.24 \\ \cline{2-6} 
 & 0.5*W & 50.982 & 72.066 & -27.07 & -22.13 \\ \hline
\end{tabular}
\caption{Results for all the transformations for ViT-L/16 (part 1 of 2)}
\label{tab:vit-l-16-1}
\end{table*}

\begin{table*}
\centering
\begin{tabular}{|l|l|r|r|r|r|}
\hline
Transformation & \begin{tabular}[c]{@{}l@{}}Degree of\\  transformation\end{tabular} & \multicolumn{1}{l|}{Top1} & \multicolumn{1}{l|}{Top5} & \multicolumn{1}{l|}{\begin{tabular}[c]{@{}l@{}}Drop in Top1 \\ accuracy \\ (absolute \%)\end{tabular}} & \multicolumn{1}{l|}{\begin{tabular}[c]{@{}l@{}}Drop in Top5 \\ accuracy \\ (absolute \%)\end{tabular}} \\ \hline
\multirow{8}{*}{\begin{tabular}[c]{@{}l@{}}Vertical translation\\ (upwards)\end{tabular}} & 0.02*H & 78.068 & 94.144 & 0.01 & -0.05 \\ \cline{2-6} 
 & 0.05*H & 77.998 & 94.018 & -0.06 & -0.18 \\ \cline{2-6} 
 & 0.1*H & 77.236 & 93.626 & -0.82 & -0.57 \\ \cline{2-6} 
 & 0.15*H & 76.57 & 93.182 & -1.48 & -1.01 \\ \cline{2-6} 
 & 0.2*H & 75.158 & 92.496 & -2.9 & -1.7 \\ \cline{2-6} 
 & 0.25*H & 73.604 & 91.218 & -4.45 & -2.98 \\ \cline{2-6} 
 & 0.35*H & 68.402 & 87.226 & -9.65 & -6.97 \\ \cline{2-6} 
 & 0.5*H & 53.064 & 74.362 & -24.99 & -19.83 \\ \hline
\multirow{8}{*}{\begin{tabular}[c]{@{}l@{}}Vertical translation\\ (downwards)\end{tabular}} & 0.02*H & 78.242 & 94.16 & 0.19 & -0.03 \\ \cline{2-6} 
 & 0.05*H & 77.79 & 94.298 & -0.26 & 0.1 \\ \cline{2-6} 
 & 0.1*H & 77.248 & 93.762 & -0.81 & -0.43 \\ \cline{2-6} 
 & 0.15*H & 76.464 & 93.322 & -1.59 & -0.87 \\ \cline{2-6} 
 & 0.2*H & 75.196 & 92.484 & -2.86 & -1.71 \\ \cline{2-6} 
 & 0.25*H & 73.446 & 91.406 & -4.61 & -2.79 \\ \cline{2-6} 
 & 0.35*H & 67.66 & 87.592 & -10.39 & -6.6 \\ \cline{2-6} 
 & 0.5*H & 52.246 & 75.276 & -25.81 & -18.92 \\ \hline
\multirow{15}{*}{\begin{tabular}[c]{@{}l@{}}Scale\\ (Zoom-in/Zoom-out)\end{tabular}} & 0.2*(H, W) & 8.664 & 18.612 & -69.39 & -75.58 \\ \cline{2-6} 
 & 0.3*(H, W) & 30.388 & 50.354 & -47.67 & -43.84 \\ \cline{2-6} 
 & 0.4*(H, W) & 49.82 & 71.978 & -28.23 & -22.22 \\ \cline{2-6} 
 & 0.5*(H, W) & 62.304 & 83.2 & -15.75 & -10.99 \\ \cline{2-6} 
 & 0.6*(H, W) & 69.278 & 88.63 & -8.78 & -5.56 \\ \cline{2-6} 
 & 0.7*(H, W) & 73.888 & 91.9 & -4.17 & -2.29 \\ \cline{2-6} 
 & 0.8*(H, W) & 75.946 & 93.104 & -2.11 & -1.09 \\ \cline{2-6} 
 & 0.9*(H, W) & 77.654 & 93.992 & -0.4 & -0.2 \\ \cline{2-6} 
 & 1.1*(H, W) & 78.062 & 94.164 & 0.01 & -0.03 \\ \cline{2-6} 
 & 1.2*(H, W) & 77.638 & 93.812 & -0.42 & -0.38 \\ \cline{2-6} 
 & 1.3*(H, W) & 77.028 & 93.478 & -1.03 & -0.72 \\ \cline{2-6} 
 & 1.4*(H, W) & 76.526 & 93.058 & -1.53 & -1.14 \\ \cline{2-6} 
 & 1.5*(H, W) & 75.658 & 92.32 & -2.4 & -1.87 \\ \cline{2-6} 
 & 1.6*(H, W) & 74.764 & 91.71 & -3.29 & -2.48 \\ \cline{2-6} 
 & 1.75*(H, W) & 73.448 & 90.706 & -4.61 & -3.49 \\ \hline
Flip left-right & - & 78.068 & 94.246 & 0.01 & 0.05 \\ \hline
Flip up-down & - & 50.664 & 72.746 & -27.39 & -21.45 \\ \hline
\end{tabular}
\caption{Results for all the transformations for ViT-L/16 (part 2 of 2)}
\label{tab:vit-l-16-2}
\end{table*}

\end{document}